\pdfoutput=1

\documentclass[11pt]{article}

\usepackage[preprint]{acl}

\usepackage{microtype}
\usepackage{graphicx}
\usepackage{subfigure}
\usepackage{booktabs} 

\usepackage{arydshln} 

\usepackage[most]{tcolorbox}
\usepackage{listings}
\usepackage{xcolor}
\definecolor{mylightblue}{RGB}{173,216,230} 
\usepackage{amsmath}
\usepackage{amssymb}
\usepackage{mathtools}
\usepackage{amsthm}
\usepackage{multirow}
\usepackage[utf8]{inputenc}
\usepackage{xspace}
\usepackage{bbm}

\usepackage[textsize=tiny]{todonotes}

\newcommand{\mname}{{\sc Pi-SQL}\xspace}

\usepackage{enumitem}

\usepackage{amsmath}
\usepackage{amsfonts}
\usepackage{algorithm} 
\usepackage{algorithmic}  
\usepackage[algo2e]{algorithm2e} 
\usepackage{pgfplots}
\pgfplotsset{compat=1.18}
\usepackage{tikz}
\usepackage{times}
\usepackage{latexsym}

\usepackage[T1]{fontenc}
\usepackage[utf8]{inputenc}
\usepackage{microtype}

\usepackage{inconsolata}

\usepackage{graphicx}

\title{\mname: Enhancing Text-to-SQL with Fine-Grained Guidance from \\Pivot Programming Languages}
\author{
    Yongdong Chi$^{1*}$, \
    Hanqing Wang$^{1}$\thanks{\ \ Equal Contribution.}, \
    Zonghan Yang$^{2}$\\
    \textbf{Jian Yang}$^{3}$,
    \textbf{Xiao Yan}$^{4}$, \
    \textbf{Yun Chen}$^{1}$, \
    \textbf{Guanhua Chen}$^{5}$ \\
    $^1$Shanghai University of Finance and Economics, $^2$Tsinghua University, \\ 
    $^3$Beihang University, 
    $^4$Wuhan University, 
    $^5$Southern University of Science and Technology \\
}

\begin{document}
\maketitle

\begin{abstract}
Text-to-SQL transforms the user queries from natural language to executable SQL programs, enabling non-experts to interact with complex databases. Existing prompt-based methods craft meticulous text guidelines and examples to facilitate SQL generation, but their accuracy is hindered by the large semantic gap between the texts and the low-resource SQL programs. In this work, we propose {\sc Pi-SQL}, which incorporates the high-resource Python program as a pivot to bridge between the natural language query and SQL program. 
In particular, {\sc Pi-SQL} first generates Python programs that provide fine-grained step-by-step guidelines in their code blocks or comments, and then produces an SQL program following the guidance of each Python program.
The final SQL program matches the reference Python program's query results and, through selection from candidates generated by different strategies, achieves superior execution speed, with a reward-based valid efficiency score up to 4.55 higher than the best-performing baseline.
Extensive experiments demonstrate the effectiveness of {\sc Pi-SQL}, which improves the execution accuracy of the best-performing baseline by up to 3.20. 
\end{abstract}

\section{Introduction} \label{sec:intro}
SQL is a standard programming language designed for managing and manipulating relational databases \cite{sqliso}. Although popular and general, SQL programs can be challenging for non-experts to write, particularly when it comes to complex data querying tasks. Text-to-SQL models convert natural language queries into executable SQL programs \citep{androutsopoulos1995natural, li2014constructing, li2024can, spider},  enabling non-experts to interact with complex databases and extract insights from big data \citep{ijcai2018-553,wang-etal-2020-rat,cao2021lgesql}. 
\begin{figure}[tbhp]
    \centering
    \includegraphics[width=1.0\columnwidth]{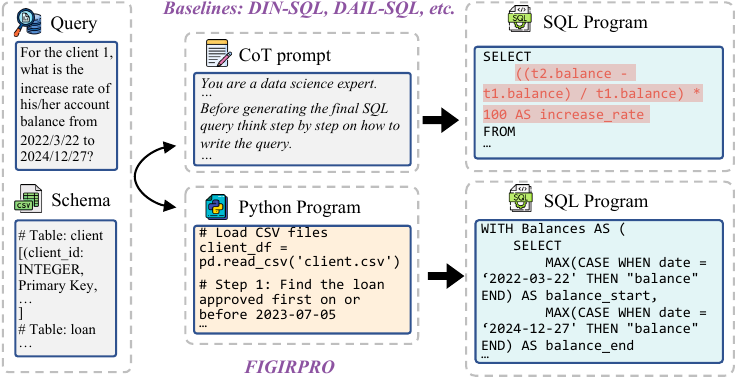}
    \caption{Given the database schema and a user query, text-to-SQL models generate an executable SQL program. Different from the text guidance produced by chain-of-thought, \mname resorts to the granular guidance from a pivot programming language.}
    \label{fig:intro}
\vspace{-15pt}
\end{figure}

Recently, many text-to-SQL models have been proposed based on large language models (LLMs), 
using either prompt engineering \cite{dinsql,tasql,c3,gao2024text,chess,chasesql} or supervised fine-tuning \cite{gao2024text,resdsql,codes,xiyan}. The prompt-based methods require meticulously crafted guidelines \cite{dinsql,tasql,c3,gao2024text,chess,chasesql} as well as curated few-shot in-domain examples \cite{dinsql,gao2024text,chess}. 
The fine-tuning-based methods rely on high-quality training data, which is expensive to obtain. Moreover, tailored to the training data domain, they may not generalize in other domains~\cite{c3,hong2024next}.

In this paper, we propose \mname, a novel prompt-based method that enhances text-to-SQL by incorporating a high-resource \underline{PI}vot programming language to provide fine-grained guidance. Motivated by multilingual pretraining \cite{xue2020mt5, lample2019cross, Huang2019UnicoderAU} and triangular machine translation \cite{kim-etal-2019-pivot,zhang-etal-2022-triangular}, \mname adopts a Python program as a pivot to bridge natural language and SQL program.
As shown in Figure~\ref{fig:intro}, compared with text-based guidance generated using chain-of-thought reasoning~\cite{dinsql,c3,gao2024text,chess}, \mname utilizes Python programs to provide more detailed step-by-step reasoning through code blocks, comments.
Existing program-of-thought-based text-to-SQL methods~\cite{xia2024r,xu2024chain} overlooked the considerable structure gap between the Python and SQL programs, 
while \mname contains approaches to mitigate the intrinsic difference between Python and SQL.

Specifically, \mname consists of an intermediate guidance preparation stage and an SQL generation stage guided by the generated Python. In the first stage,
to enhance subsequent SQL generation guided by Python, \mname employs three strategies to generate Python programs tailored for diverse SQL application scenarios, and also steers this Python generation by incorporating SQL adaptation instructions into the prompt.
In the second stage, each Python program serves as guidance for generating SQL programs. SQL programs that produce results consistent with the majority output of the Python programs are retained, and the one with the best execution efficiency is chosen. 
As a result, \mname fully leverages high-resource programming languages like Python to generate highly accurate and efficient SQL programs, without requiring few-shot examples or supervised fine-tuning with labeled data.

We compare \mname with ten state-of-the-art (SOTA) baselines on the famous BIRD~\cite {bird} and Archer~\cite{zheng-etal-2024-archer} benchmarks. The results show that \mname outperforms the baselines in both execution accuracy (EX) and reward-based valid efficiency score (R-VES), which are two popular metrics for text-to-SQL. Compared with the best-performing baseline, the EX improvement of \mname is 3.20 on the BIRD dev set.\footnote{Our code will be made publicly available.}

\begin{figure*}[tbhp]
    \centering
    \includegraphics[width=0.9\textwidth]{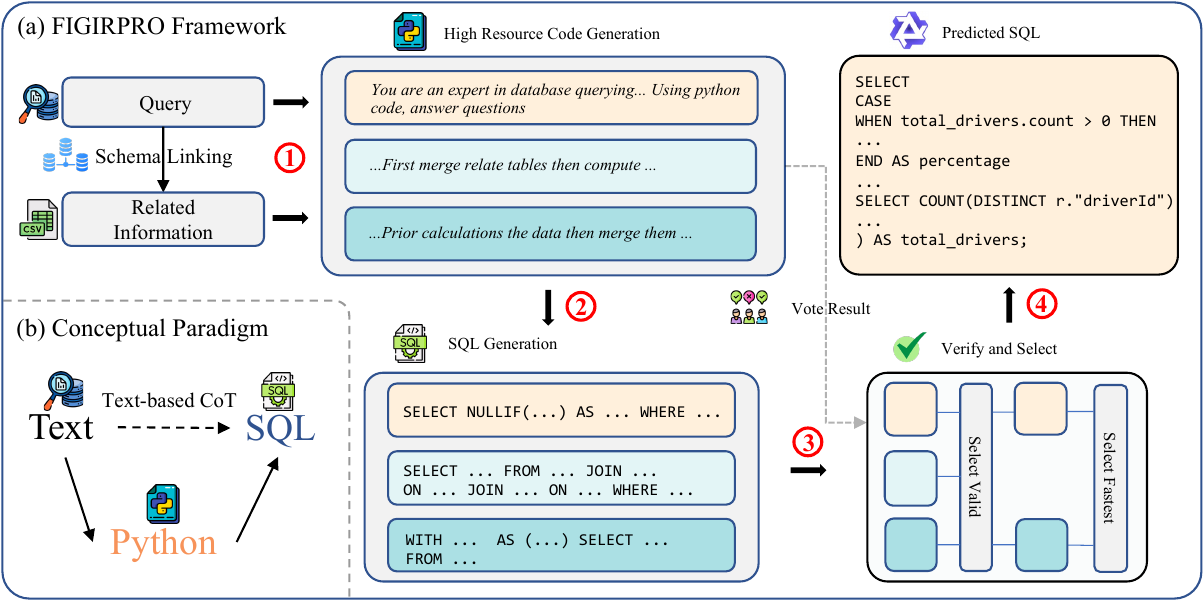}
    \caption{Overview of our \mname method. (a) The workflow of \mname. It incorporates high-resource programming languages like Python to provide step-by-step fine-grained guidance and verification to enhance LLM-based text-to-SQL. (b) The difference between \mname and existing text-based CoT approaches.}
    \label{fig:overview}
\vspace{-10pt}
\end{figure*}

\section{Related Work} \label{sec:related_work}

\paragraph{Prompt-Based Text-to-SQL Methods}

Given the strong generalization ability of LLMs, recent mainstream research has shifted towards leveraging their powerful few-shot and zero-shot capabilities through prompt-based approaches. 
Numerous research efforts have focused on enhancing text-to-SQL performance from various aspects, including improved schema linking~\citep{dinsql,gao2024text,c3}, the selection of more effective few-shot demonstrations~\citep{dinsql,sun2023sql,chen2023teaching,gao2024text,tasql,chess,chasesql}, incorporating natural language chain-of-thought reasoning~\citep{dinsql,gao2024text,chess,chasesql}, and using the self-consistency method to boost performance with additional test-time computing~\citep{sun2023sql,gao2024text,lee2024mcs,chess,maamari2024the,chasesql}. Different from these works, \mname achieves better text-to-SQL performance by leveraging the fine-grained guidance from high-resource programming languages in a zero-shot setting. 
\paragraph{Program of Thoughts}
Program of Thoughts (PoT) is an extension of the Chain of Thought (CoT) prompting strategy, aiming to mitigate errors in intermediate reasoning by executing intermediate steps as Python programs. 
PoT has been widely adopted in various tasks~\citep{payoungkhamdee2025towards,sahu-etal-2024-pelican,sarch2024vlm,zhang-etal-2024-tinychart}, due to its improved reliability of numerical and logical inference. 
However, its application to text-to-SQL remains limited. $R^3$~\citep{xia2024r} directly employs PoT to generate Python code before SQL to guide the SQL generation, while~\citet{xu2024chain} incorporates PoT into the text-to-SQL training process. These approaches overlook the semantic and structural gap between Python and SQL.
In contrast, \mname's design provides better SQL generation guidance with Python, which mitigates the intrinsic difference between Python and SQL.

\section{Method} \label{sec:method}

\subsection{Motivation and Insights} \label{sec:overview}
Previous works \cite{chasesql,chess,dinsql} apply text-based reasoning as the guidance to generate SQL responses. 
However, they still struggle with various hard-level queries and database schemas, primarily due to the limited SQL corpus encountered during pretraining.
Unlike these approaches, we propose to leverage fine-grained guidance from high-resource programming languages (\mname). High-resource programming languages such as Python serve as pivot languages, bridging the gap between SQL and natural language, akin to the triangular neural machine translation model \cite{zhang-etal-2022-triangular}.

The \mname framework is motivated by the advancements achieved through multilingual pretraining \citep{xue2020mt5, lample2019cross, Huang2019UnicoderAU} and triangular neural machine translation \cite{kim-etal-2019-pivot,zhang-etal-2022-triangular}. (1) The low-resource languages (LRL) share similar syntaxes and lexemes with the high-resource languages from the same language family. During multilingual pretraining, the shared information serves as anchor points to better align the representation space of these languages, improving the performance on LRL via cross-lingual transfer.
(2) In the case of triangular machine translation, a high-resource pivot language is incorporated to improve the translation from the source language to the target language. For example, instead of directly translating from English to Estonian, Finnish is used as a pivot language, as it belongs to the same language family as Estonian, and the translation from Finnish to Estonian is easier. The English text is first translated into Finnish, and then the Finnish translation is further translated into Estonian. The triangular translation improves the performance of low-resource language pairs.

The situation is similar for SQL and its corresponding high-resource programming language, Python. Both programs are widely applied in the field of data analysis and share similar logic and keywords. During code pretraining, SQL aligns with the representation space of Python and benefits from the large-scale Python data, similar to the case in multilingual pretraining. Meanwhile, Python and natural language are well aligned in a shared representation space as both are high-resource and are jointly pretrained with shared anchor points like comments in Python code. 

In this way, the \mname invites Python as a pivot language to bridge user queries in natural language and the low-resource SQL programs for better text-to-SQL performance, as shown in Figure~\ref{fig:overview}. 

When compared with direct SQL generation, guidance with the corresponding Python program has several advantages, illustrated as follows. 
\vspace{-8pt}
\begin{itemize}[left=0pt, itemsep=-2pt]
    \item \textbf{Proficient}. Large-scale Python data contributes to the proficiency and accuracy of data analysis with Python for LLMs. 
    \item \looseness=-1 \textbf{Fine-Grained}. Different from the nested operations in SQL, Python programs decompose complex data query tasks into verifiable code blocks as well as comments to form step-by-step reasoning trajectories. The execution results, such as exceptions or errors, also serve as fine-grained feedback and self-reminders for SQL generation.
    \item \textbf{Modular}. Various Python packages abstract the reasoning process and facilitate the generation of corresponding code blocks.
\end{itemize}
Specifically, given a complex user query $q_u$, the LLM first responds with a high-resource programming language $C^p_i$ with different data analysis strategies $G_i$. The LLM then generates the SQL program $C^s_i$ with the guidance of the Python program $C^p_i$ and its execution results $E^p_i$. \mname consists of two stages: Intermediate Guidance Preparation (Section~\ref{sec:igp_sec}) and SQL Generation (Section~\ref{sec:sql_sec}).

\subsection{Intermediate Guidance Preparation} \label{sec:igp_sec}
The Intermediate Guidance Preparation stage aims to generate diverse fine-grained Python programs as guidance for text-to-SQL tasks. Following previous works \citep{chess,esql,cao2024rsl}, we incorporate a schema-linking module to retrieve relevant tables and columns from a schema with the user query. The retrieved data is further converted to \texttt{csv} files for interaction with the Python program.\footnote{We measured that constructing the \texttt{csv} files takes only about 0.024 seconds per question on average, which is negligible compared to the inference time.} 
However, the transformation from Python to SQL cannot be directly applied due to the fundamental differences between Python and SQL. While Python is a procedural language, SQL is a declarative, structured query language. This distinction may pose challenges in using Python to guide SQL generation. The result of $R^3$ in Table~\ref{tab:main-results} demonstrates this. To bridge this gap, we introduce SQL-specific strategies (see Section~\ref{sec:igp_sec}) as shown in Algorithm~\ref{alg:framework} along with a set of Python-to-SQL adaptation rules. These components help mitigate the mismatch between the two languages and allow us to more effectively leverage Python’s execution signals in the SQL selection process. 

\paragraph{Diverse Python Generation Strategies.}
The \mname framework crafts three different strategies to guide the reasoning trajectories of Python program generation. As a high-level, interpreted scripting programming language, Python has different reasoning paths compared to SQL in the data analysis task. The Python program analyzes data sequentially, where the relevant data is first filtered and then combined for further analysis. The SQL program benefits from the efficient data analysis engine that SQL users are accustomed to first combining all relevant data and then further analyzing. 
To better generate Python programs that can guide SQL generation across different application scenarios, we design three distinct strategies: merge, filter, and direct.

These strategies are illustrated below, encouraging intermediate guidance to incorporate diverse reasoning trajectories:
\begin{itemize}[left=0pt, itemsep=-2pt]
    \item \textbf{Merge-First Strategy}. This strategy asks the LLM to merge and join the relevant columns first based on the input information. Then filter and extract the required data. This strategy aligns with the design philosophy of relational databases. As different data are decomposed and stored separately, they are first reconstructed with the foreign keys and then analyzed.
    \item \textbf{Filter-First Strategy}. This prompt guideline suggests the LLM filter and prepare the relevant columns first based on the input information. Then the model directly generates further analysis code based on the filtered data. This strategy follows the vanilla practice of Python programs in data analysis tasks.
    \item \textbf{Vanilla Direct Generation}. This strategy does not impose any suggestion on the LLM. The model generates the Python program in a freestyle learned during pretraining. 
\end{itemize}
\looseness=-1 With the guidance of different strategies, the LLM is expected to generate Python programs with different reasoning trajectories. 
Moreover, Python offers a richer set of libraries and functions for data analysis that are not supported by SQL engines. To enhance Python-to-SQL adaptation and improve the quality of guidance, we also explicitly prompt LLMs to use APIs and functions that closely resemble valid SQL operations.

\paragraph{Verification of Python Program.}
The diverse generations of Python programs are verified with execution with the \texttt{csv} files.\footnote{We discuss the execution time in Appendix~\ref{app:execution_time}.} The self-consistency method \cite{wang2023selfconsistency} is employed to determine the reference query result for the user's query from all Python execution outcomes. This reference result is then used to select the final SQL response in a subsequent stage. We contend that selecting the SQL based on Python results effectively serves as a double-check mechanism, further ensuring the faithfulness of the chosen SQL.

\begin{table*}[thbp]
\vspace{-10pt}
\centering
\resizebox{0.9\textwidth}{!}{

\begin{tabular}{lccccccccccccc}
\toprule
\multirow{2}{*}{Method} & \multirow{2}{*}{Zero-shot} & \multicolumn{2}{c}{Few-shot} & \multicolumn{2}{c}{Archer} & \multicolumn{2}{c}{BIRD} & \multicolumn{2}{c}{B-Simple} & \multicolumn{2}{c}{B-Moderate} & \multicolumn{2}{c}{B-Challenging} \\
\cmidrule(lr){3-4} \cmidrule(lr){5-6} \cmidrule(lr){7-8} \cmidrule(lr){9-10} \cmidrule(lr){11-12} \cmidrule(lr){13-14}
& & Fixed & Dynamic & EX & R-VES & EX & R-VES & EX & R-VES & EX & R-VES & EX & R-VES \\
\midrule
Vanilla & $\checkmark$ & & & 10.58 & 10.19 & 54.30 & 55.94 & 61.84 & \underline{64.96} & 43.10 & 42.16 & 42.07 & 42.55\\
C3 & $\checkmark$ & & & 16.35 & 23.04 & 57.37 & 53.65 & 65.51 & 61.40 & 46.98 & 43.66 & 38.62 & 36.19 \\
DIN-SQL & & & $\checkmark$ & 8.65 & 13.56 & 50.07 & 46.80 & 58.16 & 54.29 & 39.44 & 37.08& 32.41 & 30.10 \\
DAIL-SQL & & & $\checkmark$ & 16.35 & 18.07 & 55.02 & 51.02 & 62.16 & 57.67 & 46.98 & 43.40 & 35.17 & 33.04\\
TA-SQL & & $\checkmark$ & & 8.65 & 10.06 & 55.15 & 52.06 & 63.35 & 59.79 & 44.18 & 41.33 & 37.93 & 37.08 \\
R\textsuperscript{3} & & &$\checkmark$ &20.19 &21.27& 52.67 & 47.72 & 57.95 & 53.23 & 44.83 & 39.87 & 44.14 & 37.71 \\
CHESS & & $\checkmark$ & & 21.15 & 25.81 & 61.02 & 56.91 & \underline{68.54} & 64.20 & 49.78& 46.15 & \underline{48.97} & 44.81 \\
CHASE-SQL & & &$\checkmark$ & \textbf{25.96}& \underline{28.70} & \underline{61.34} & \underline{59.16} & \underline{68.54} & 64.35 & \underline{52.80} & \underline{52.11} & \underline{48.97} & \underline{48.69} \\
E-SQL & & $\checkmark$ & &16.35 &17.13 & 58.47 & 54.80 & 65.08 & 61.05 & 51.29 & 47.89 & 39.31 & 37.00 \\
RSL-SQL & & & $\checkmark$ &13.46 &14.87 & \underline{61.34} & 56.76 & 67.89 & 63.13 & \underline{52.80} & 48.38 & 44.18 & 42.96 \\
\midrule
\mname & $\checkmark$ & & & \underline{25.00} & \textbf{30.10} & \textbf{64.54}& \textbf{63.71}& \textbf{70.92}& \textbf{70.06}& \textbf{56.47}& \textbf{55.63}& \textbf{49.66}& \textbf{49.06} \\
$\Delta$ & - & - & - & -0.96 & +1.40 & +3.20 & +4.55 & +2.38 & +5.10 & +3.67 & +3.52 & +0.69 & +0.37 \\
\bottomrule
\end{tabular}}
\caption{
Execution Accuracy (EX) and Valid Efficiency Score (VES) on the Archer and BIRD datasets. BIRD provides the results for different query difficulty levels. We indicate whether a method is zero-shot or few-shot, and if few-shot, whether it uses fixed or dynamic shots. 
The best and runner-up results for each case are marked with  \textbf{bold} and  \underline{underline}, respectively.  `$\Delta$' is the performance gain of \mname over the best-performing baseline.}
\label{tab:main-results}
\vspace{-10pt}
\end{table*}

\subsection{SQL Generation with Python Guidance} \label{sec:sql_sec}

In the second stage, the SQL responses are generated with guidance from the corresponding fine-grained Python program, as shown in Algorithm~\ref{alg:framework}. 
Subsequently, the \mname framework verifies these generated SQL programs by executing them in the database. An SQL program is deemed a valid candidate if its execution result matches the reference query result. Finally, among all valid candidates, the highest execution efficiency is selected as the final SQL response, $R^s_f$. This entire stage, including the SQL generation process and the final selection, is detailed in Algorithm~\ref{alg:framework}.

\SetKwComment{Comment}{// }{}
\vspace{-8pt}
\begin{algorithm}
\DontPrintSemicolon
  \footnotesize
  \KwData{LLM $\theta$, user query $q_u$, relational database $D$ associated with the query, and strategy of Python program generation $C$.}
  \KwResult{Model predicted SQL program $R^s_f$}
  $\mathcal{R}_{p} = \mathrm{GeneratePython}(\theta, C, q_u, D)$ \\
  $\mathcal{R}_{s} = \emptyset$ \\
  \For{each Python program $R^p_i$ in $\mathcal{R}_{p}$}{ 
      $E^p_i = \mathrm{Execute}(R^p_i)$  \Comment{Execute the code and get results}
      $R^s_i = \mathrm{GenerateSQL}(\theta, q_u, D, R^p_i)$  \Comment{Generate SQL for the code} 
      $\mathcal{R}_{s} = \mathcal{R}_{s} \cup {R^s_i}$ 
  }
  $\mathcal{R}_{se} = \emptyset$ \\
  \For{each SQL program $R^s_i$ in $\mathcal{R}_{s}$}{ 
      $E^s_i = \mathrm{ExecuteSQL}(R^s_i, D)$ \Comment{Execute SQL and get database results}
      $\mathcal{R}_{se} = \mathcal{R}_{se} \cup \{(R^s_i, E^s_i)\}$ 
  }
  $\mathrm{MajorityResult} = \mathrm{FindMostFrequent}(\mathcal{R}_{p})$  \Comment{Find the most frequent execution result among Python codes}
  $\mathrm{ValidSQL} = \mathrm{SelectValidSQL}(\mathcal{R}_{se}, \mathrm{MajorityResult}, D)$ \Comment{Select valid SQL that matches majority result}
  
  $R^s_f = \arg\min_{r \in \mathrm{ValidSQL}} \mathrm{ExecutionTime}(r)$ \Comment{Select SQL with the least execution time}
  \Return $R^s_f$
  \caption{Algorithm for \mname framework}
  \label{alg:framework}
\end{algorithm}
\vspace{-15pt}

\section{Experiments} \label{sec:experiments}

\subsection{Experiment Settings}  \label{sec:exp_setup}

\paragraph{Benchmarks}
We conduct experiments on two widely recognized text-to-SQL datasets: BIRD \cite{bird} and Archer \citep{zheng-etal-2024-archer}. They are designed to encompass various real-world scenarios, featuring simple and complex query structures. Spider~\citep{spider} is not selected for evaluation for the following reasons: 1) Our method focuses on more challenging text-to-SQL queries, whereas the queries in Spider are relatively easy; 2) We have found that the ground-truth SQL programs in Spider are noisy \cite{zhong-etal-2023-non}, which makes the evaluation results unreliable. More details about the benchmarks are available in Appendix~\ref{sec:app_benchmark}.

\paragraph{Metrics} We utilize Execution Accuracy (EX)~\citep{spider} and Reward-based Valid Efficiency Score (R-VES)~\citep{bird} as evaluation metrics to assess the methods' performance.

\noindent $\bullet$ \textbf{Execution Accuracy (EX):} EX measures the ratio of correctly predicted SQL programs by comparing their execution results with those of the ground-truth SQL programs on the same database instance.

\noindent $\bullet$ \textbf{Reward-based Valid Efficiency Score (R-VES):} 
R-VES evaluates the performance of models that generate SQL queries, considering their accuracy and runtime performance. As an improvement over the previous Valid Efficiency Score (VES), R-VES incorporates the execution time of correct queries into the evaluation while mitigating the influence of abnormal or outlier execution times.

\paragraph{Implementation Details}
To ensure a fair comparison in a unified setting, we use the same LLM backbone with a temperature of 0 and a maximum token limit of 4096 for \mname and all baselines. We choose GPT-4o-mini~\citep{openai2024gpt4ominidocs} as the backbone for the computational constraints.\footnote{For example, performing CHESS on the BIRD dev set with GPT-4o~\citep{hurst2024gpt} costs approximately \$ 800.} And to compare with advanced performance, we also perform \mname with Qwen2.5-Coder-32B-Instruct~\citep{hui2024qwen2} in section~\ref{sec:adv_result}. 
We adopt the schema linking module from RSL-SQL~\cite{cao2024rsl}, as schema linking is not the primary focus of this work.
Furthermore, we evaluate \mname using various backbone models in Section~\ref{sec:diff_backbone} and on models of varying scales in Appendix~\ref{app:scale}.

\paragraph{Baselines}

We compare \mname with ten baselines: \textbf{Vanilla}, \textbf{C3}~\citep{c3}, \textbf{DIN-SQL}~\citep{dinsql}, \textbf{DAIL-SQL}~\citep{gao2024text}, \textbf{TA-SQL}~\citep{tasql}, $\mathbf{R}^{\mathbf{3}}
$~\citep{xia2024r} \textbf{CHESS}~\citep{chess}, \textbf{CHASE-SQL}~\citep{chasesql}, \textbf{E-SQL}~\citep{esql}, and \textbf{RSL-SQL}~\citep{cao2024rsl}. The vanilla method is the same as \mname, except for the absence of Python guidance. We introduce each baseline in detail in the Appendix~\ref{sec:app_baselines}.

\subsection{Main Results} \label{sec:main}

Table~\ref{tab:main-results} shows the comparison results of \mname against the baselines. Overall, \mname outperforms all other baselines on both BIRD and Archer in terms of execution accuracy and efficiency. This is impressive because our zero-shot approach outperforms the zero-shot baseline vanilla, C3, and the few-shot baselines, specifically DIN-SQL, DAIL-SQL, TA-SQL, and CHESS.

When comparing \mname with baselines across different query difficulty levels on BIRD, we find that
\mname shows consistent improvements over the baselines on different difficulty levels. Specifically, it improves over baselines by 2.38 to 12.97 EX on simple-level queries, 3.67 to 17.03 EX on moderate queries, and 0.69 to 17.25 EX on challenging queries. 
This may be attributed to the rich data processing capabilities of Python, which enable large language models (LLMs) to handle a wide range of queries more effectively. \mname also achieves consistent improvements on the R-VES benchmark over the baselines by 4.55 to 16.91, further validating its effectiveness. These results highlight the strong potential of our method in real-world scenarios, as it: (1) can handle queries of varying difficulty across diverse contexts, and (2) generates SQL queries that are both accurate and efficient.

Table \ref{tab:token_usage} and \ref{tab:token_cost} in Appendix \ref{app:cost_compare} compare the inference token usage and inference cost of \mname with the baselines. \mname has a lower inference cost than the best-performing baseline CHESS, a higher cost than vanilla and C3, and a comparable cost to the other baselines.
We believe this test-time cost is justifiable for two reasons: 1) It substantially enhances performance over vanilla and C3, especially for challenging queries; 2) The generated SQL programs can be executed by users multiple times in practice, making the significant R-VES improvement achieved by \mname particularly valuable. In Section \ref{sec:abs}, we provide additional evidence that the improvement of \mname over the vanilla method is not solely due to the increased test time computing.

\subsection{Ablation Study} \label{sec:abs}

In this section, we conduct an ablation study on 3 key components of \mname using the BIRD benchmark: the Python generation strategy, the SQL selection method, and the Python-SQL adaptation. For the Python generation strategy, we evaluate four variants: using merge, filter, or direct individually, or using a combination of all three. For the SQL selection method, we either select by referring to the Python execution result (cross-verification) or by taking a majority vote from the SQL execution results. Regarding the Python-SQL adaptation, we compare the performance of \mname with and without this adaptation. We also add a vanilla+self-consistency baseline, for which we directly generate $N$ SQLs for each query and select the final SQL program using self-consistency of the SQL execution results. We set the value of $N$ to 11 to ensure that the token cost of this baseline matches that of \mname, and the temperature to 0.5 is determined on a validation set.

\begin{table}[tp]
\centering
\resizebox{0.48\textwidth}{!}{%
\begin{tabular}{lcccccccc}
\toprule
\multirow{2}{*}{Setup}& \multicolumn{2}{c}{Overall} & \multicolumn{2}{c}{Simple} & \multicolumn{2}{c}{Moderate} & \multicolumn{2}{c}{Challenging}  \\
\cmidrule(lr){2-3} \cmidrule(lr){4-5} \cmidrule(lr){6-7} \cmidrule(lr){8-9}
 & EX & R-VES & EX & R-VES & EX & R-VES & EX & R-VES  \\ \midrule
 Vanilla & 54.30&55.94&61.84&64.96&43.10&42.16&42.07&42.55 \\
 Vanilla+SC  & 59.71&57.77&66.27&64.19&52.37&50.99&41.38&38.57 \\
 \midrule
\multicolumn{9}{c}{\textit{Ablation on code generation mode}} \\
\midrule
Merge &61.93&59.83&68.11&66.08&53.45&51.34&49.66&47.13 \\							
Filter &61.99&60.19&67.89&66.01&54.09&52.48&49.66&47.74 \\
Direct &62.58&61.09&68.97&67.41&53.88&52.69&49.66&47.67 \\
Ours(Mixed) &\textbf{64.54}& \textbf{63.71}& \textbf{70.92}& \textbf{70.06}& \textbf{56.47}& \textbf{55.63}& \textbf{49.66}& \textbf{49.06} 
\\\midrule
\multicolumn{9}{c}{\textit{Ablation on SQL selection method}}  \\
\midrule
Mixed+SC &63.62&61.53&70.05&67.88&54.74&52.77&51.03&49.09\\
Ours(Mixed+CV)& \textbf{64.54}& \textbf{63.71}& \textbf{70.92}& \textbf{70.06}& \textbf{56.47}& \textbf{55.63}& \textbf{49.66}& \textbf{49.06}
\\\midrule
\multicolumn{9}{c}{\textit{Ablation on Python-SQL adaptation}}  \\
\midrule
W/O adaptation &63.36&61.97&70.16&68.83&54.31&52.86&48.97&47.34 \\
Ours &\textbf{64.54}& \textbf{63.71}& \textbf{70.92}& \textbf{70.06}& \textbf{56.47}& \textbf{55.63}& \textbf{49.66}& \textbf{49.06} \\																			
\bottomrule
\end{tabular}}
\caption{Ablation study on BIRD dataset. We perform ablation for the Python generation strategy and SQL selection method. `SC' means self-consistency while `CV' means cross verification. The best result for each case is highlighted in \textbf{bold}.}
\label{tab:ablation}
\end{table}

Table \ref{tab:ablation} shows the ablation results. When using the same inference token cost, \mname outperforms the vanilla method by 4.83 EX and 5.94 R-VES. This indicates that our performance improvement is not solely attributable to the test time scaling law but rather to the effective guidance provided by high-resource programming languages.

For ablation on Python generation strategies, we observe that using any single Python generation strategy substantially improves upon vanilla GPT-4o-mini, achieving an improvement over 7.63 EX and 3.89 R-VES. However, mixing all strategies yields the best performance across all difficulty levels, surpassing the best single method by 1.96 EX and 2.62 R-VES. This indicates that different strategies are complementary, highlighting the importance of using a mixed approach.

Our cross-verification approach for SQL selection consistently outperforms the self-consistency method across all difficulty levels and evaluation metrics. This could be attributed to the diversity of errors made by Python and SQL, which contrasts with the more similar errors produced by different SQLs.

Python-SQL adaptation consistently improves the performance of \mname, achieving an overall gain of 1.18 in EX score and 1.74 in R-VES, with particularly notable improvements on moderate and challenging queries. This may be attributed to the fact that more complex problems benefit from clearer and more relevant guidance. 

Consequently, the mixed generation strategies, cross verification method, and Python-SQL adaptation collectively enhance the generation quality of \mname, distinguishing it from previous, direct PoT-based methods such as $R^3$.

\section{Analyses}

\subsection{Comparison with SOTA Results} \label{sec:adv_result}
Due to computational constraints, 
We could not perform a direct comparison with other methods using SOTA LLMs such as GPT-4o under the same experimental setting. Instead, we evaluate \mname with Qwen2.5-Coder-32B-Instruct and compare its performance against the SOTA methods such as Distillery~\citep{maamari2024the}, OpenSearch-SQL~\citep{xie2025opensearch}, and XiYan-SQL~\citep{xiyan} reported on the BIRD leaderboard\footnote{\url{https://bird-bench.github.io}} in Table~\ref{tab:sota_result}.
The results show that \mname, as a zero-shot approach, can achieve performance comparable to or even surpassing SOTA methods that rely on fine-tuning, refinement, or powerful proprietary models, using only a 32B open-source model. 
As shown in Appendix~\ref{sec:app_simple_refinement}, a naive refinement method enhances \mname by 0.58 in EX score and 3.14 in R-VES score, underscoring its compatibility with refinement strategies.

\begin{table}[ht]
\centering
\resizebox{0.48\textwidth}{!}{%
\begin{tabular}{lcccc}
\toprule
Method & Backbone & Finetuned & With Refinement & EX  \\
\midrule
R\textsuperscript{3} & GPT-4        & & \checkmark  & 61.80 \\
CHESS$^\dagger$          & GPT-4o           & & \checkmark & 65.00 \\
E-SQL$^\dagger$           & GPT-4o           &  & \checkmark & 65.58 \\
Distillery$^\dagger$      & GPT-4o           & \checkmark & & 67.21\\
OpenSearch-SQL$^\dagger$  & GPT-4o           &  & \checkmark & 69.30\\
CHASE-SQL$^\dagger$       & Gemini 1.5 pro       & \checkmark & \checkmark & 73.01\\
XiYan-SQL$^\dagger$       & UNK           & \checkmark & \checkmark & \textbf{73.34}\\
\midrule
XiYan-SQL$^\dagger$       & Qwen2.5-Coder-32B           & \checkmark & \checkmark & 67.01\\
Ours   & Qwen2.5-Coder-32B       &  & &  \textbf{67.40} \\
\bottomrule
\end{tabular}}
\caption{Comparison with state-of-the-art (SOTA) methods on the BIRD dev set. Due to computational cost, \mname was evaluated using a 32B open-source model. $^\dagger$Results are taken from the BIRD leaderboard. $R^3$ results are from~\citep{xia2024r}.}
\label{tab:sota_result}
\end{table}

\subsection{Comparison with Zero-shot Methods}

To ensure fair comparison with the zero-shot \mname, all baselines were evaluated in a zero-shot setting. Although methods like DIN-SQL are not zero-shot, we include them to assess performance changes without in-context examples, thereby revealing the dependency of such methods on few-shot demonstrations. Table~\ref{tab:zero-shot} shows \mname significantly outperforms baselines under these conditions, achieving 5.09-27.97 higher EX and 3.97-29.26 higher R-VES scores. This underscores \mname's advantages: no need for complex shot design and superior generalization. While most few-shot baselines decline without examples, DIN-SQL, DAIL-SQL, RSL-SQL, and CHASE-SQL show relative stability. In contrast, TA-SQL, CHESS, $R^3$, and E-SQL experience substantial drops (EX score reductions of 18.58, 24.00, 4.10, and 4.23, respectively), as their rule-based SQL generation heavily relies on few-shot examples for effective LLM rule interpretation and application.

\begin{table}[t]
\centering
\resizebox{0.48\textwidth}{!}{
\begin{tabular}{lcccccccccc}
\toprule
\multirow{2}{*}{Method} & \multicolumn{4}{c}{Overall} & \multicolumn{2}{c}{Simple} & \multicolumn{2}{c}{Moderate} & \multicolumn{2}{c}{Challenging} \\
\cmidrule(lr){2-5} \cmidrule(lr){6-7} \cmidrule(lr){8-9} \cmidrule(lr){10-11}
& EX & R-VES & $\Delta$EX & $\Delta$R-VEX & EX & R-VES & EX & R-VES & EX & R-VES \\
\midrule
Vanilla &54.30&55.94&-&-&61.84&64.96&43.10&42.16&42.07&42.55 \\
C3 & 57.37 & 53.65 & - & - & 65.51 & 61.40 & 46.98 &43.66 & 38.62 & 36.19 \\
DIN-SQL & 50.85 & 47.51 & +0.78 & +0.71 & 58.70 & 55.04 & 41.16 &38.44&31.72 &28.45 \\
DAIL-SQL & 53.45 & 48.80 & -1.57 & -2.22 &59.56 & 54.88 & 47.41 & 44.86 & 33.79 & 33.49\\
TA-SQL & 36.57 &34.45 & -18.58 & -17.61 & 42.92 &40.43 &28.02 &26.23 &23.45&22.60\\
CHESS & 37.02 & 37.46 & -24.00 & -19.45 & 44.75 & 45.19 & 26.07 & 26.33 & 22.75 &23.79 \\
R\textsuperscript{3} & 48.57 & 49.13 & -4.10 & +1.41 & 54.05 & 54.33 & 42.67 & 43.89 & 32.41 &32.77 \\
RSL-SQL & 59.45 & 58.95 & -1.89 & +2.19 & 66.27 & 66.52 & 51.08 & 49.76 & 42.76 &40.11\\
E-SQL & 54.24 & 49.13 & -4.23 & -5.67 & 60.91 & 61.26 & 50.00 & 50.49 & 34.78 &33.64 \\
CHASE-SQL & 59.25 & 59.74 & -0.79 & +0.58 & 63.67 & 64.72 & 53.66 &53.29 & 48.96 &48.62 \\
\midrule
\mname &\textbf{64.54}& \textbf{63.71}& - & - & \textbf{70.92}& \textbf{70.06}& \textbf{56.47}& \textbf{55.63}& \textbf{49.66}& \textbf{49.06} \\
\bottomrule
\end{tabular}}
\caption{Comparing \mname with zero-shot baselines on BIRD dataset. $\Delta$EX/VES denotes the zero-shot EX/R-VES minus the few-shot EX/R-VES. The best result for each case is highlighted in \textbf{bold}.}
\label{tab:zero-shot}
\end{table}

\subsection{Using Different LLM Backbones}
\label{sec:diff_backbone}

In this section, we further investigate the performance of \mname on open-sourced LLM backbones, including Qwen2.5-Coder-32B-Instruct, QwQ-32B~\citep{qwq32b}, and Gemma-3-27B-IT~\citep{gemma_2025}. 

\begin{table}[t]
\centering
\resizebox{1.0\columnwidth}{!}{%
\begin{tabular}{lcccccccc}
\toprule
\multirow{2}{*}{Model} & \multicolumn{2}{c}{Overall} & \multicolumn{2}{c}{Simple} & \multicolumn{2}{c}{Moderate} & \multicolumn{2}{c}{Challenging} \\
\cmidrule(lr){2-3} \cmidrule(lr){4-5} \cmidrule(lr){6-7} \cmidrule(lr){8-9}
 & EX & R-VES & EX & R-VES & EX & R-VES & EX & R-VES \\
 \midrule
\multicolumn{9}{l}{\textit{Vanilla method}} \\
\midrule
Qwen-Coder &59.97&58.61&64.76&63.64&55.17&53.40&44.83&43.12 \\
QwQ & 55.08&54.06&62.49&61.78&46.34&44.98&35.86&33.87\\
Gemma3 & 58.87&57.51&64.32&62.96&51.72&50.71&46.90&44.49 \\
GPT-4o-mini &54.30&55.94&61.84&64.96&43.10&42.16&42.07&42.55 \\
\midrule
\multicolumn{9}{l}{\textit{With \mname}} \\
\midrule
Qwen-Coder & \textbf{67.40} & 65.54 & \textbf{72.86} & 71.14 & \textbf{59.91} & \textbf{57.89} & 56.55 & 54.29 \\
QwQ & 64.34&62.79&71.14&69.41&56.90&55.52&44.83&43.80 \\
Gemma3  & 66.42&\textbf{65.62}&\textbf{72.86}&\textbf{72.08}&56.25&55.63&\textbf{57.93}&\textbf{56.41} \\
GPT-4o-mini &64.54& 63.71& 70.92& 70.06& 56.47& 55.63& 49.66& 49.06
 \\
\bottomrule
\end{tabular}}
\caption{Performance of \mname with different LLM backbones on the BIRD dev set. The best result for each case is highlighted in \textbf{bold}.}
\label{tab:model-performance}
\vspace{-10pt}
\end{table}

As shown in Table~\ref{tab:model-performance}, \mname consistently enhances the performance of vanilla methods across all backbones, irrespective of their architectures and model types. This demonstrates the versatility and robustness of our method, which requires no additional training and can be applied to a wide range of backbones. To further demonstrate the effectiveness of \mname, we compare it with fine-tuned methods in Appendix~\ref{app:compare_finetuned}, perform \mname on the Qwen-Coder-Instruct series with different scales in Appendix~\ref{app:scale}, and present case studies in Appendix~\ref{app:case_code_generation} and Appendix~\ref{app:casestudy}.

\subsection{Analyzing Python Generation Results} \label{sec:code}

\begin{table}[t]
\centering
\resizebox{1.0\columnwidth}{!}{%
\begin{tabular}{lcccccc}
\toprule
\multirow{2}{*}{Model} & \multicolumn{3}{c}{Overall} & \multicolumn{3}{c}{Challenging} \\
\cmidrule(lr){2-4} \cmidrule(lr){5-7}
 & VanS & {\sc Pi}S & {\sc Pi}P & VanS & {\sc Pi}S & {\sc Pi}P \\
\midrule
Qwen-Coder   & 59.97 & 67.40 & 66.42 & 44.82 & 56.55 & 61.37 \\
QwQ  & 55.08 & 64.34 & 65.58 & 35.86 & 44.82 & 62.06 \\
Gemma3  & 58.86 & 66.42 & 62.12 & 46.89 & 57.93 & 54.48 \\
GPT-4o-mini & 54.30 & 64.54 & 65.44 & 42.07  & 49.66 & 59.31 \\
\bottomrule
\end{tabular}}
\caption{Performance (EX) of Python program on the BIRD dev set. VanS and {\sc Pi}S denote SQL performance without and with Python guidance, while {\sc Pi}P denotes Python performance.}
\label{tab:code-performance}
\vspace{-10pt}
\end{table}

\begin{table}[t]
\centering
\resizebox{1.0\columnwidth}{!}{%
\begin{tabular}{lcccccc}
\toprule
\multirow{2}{*}{Model} & \multicolumn{3}{c}{Overall} & \multicolumn{3}{c}{Challenging} \\
\cmidrule(lr){2-4} \cmidrule(lr){5-7}
 & Merge & Filter & Direct & Merge & Filter & Direct \\
\midrule
Qwen-Coder  & 34.49&35.27&30.25& 31.72&33.10&35.17 \\
QwQ & 34.49&31.42&34.09 & 33.10&22.07&44.83 \\
Gemma3  & 31.29&33.64&35.07 & 29.66&33.79&36.55 \\
GPT-4o-mini & 31.94 & 34.15 & 33.89 & 35.17& 31.03 & 33.79 \\
\bottomrule
\end{tabular}}
\caption{Distribution of Python selection rates across different Python generation strategies.}
\label{tab:method-distribution}
\vspace{-10pt}
\end{table}

To better understand how the pivot program improves SQL generation, we analyze the intermediate results from Python. Table~\ref{tab:code-performance} shows the EX of the final Python result, selected with self-consistency after executing all Python candidate programs. We can conclude that: 1) PIS could significantly outperform VanS, demonstrating the effectiveness of Python guidance. 2) Second, we observe that PiP shows advantages over PIS on the challenging subsets. This suggests that there is still room for improvement in our method, and using Python as guidance for SQL generation remains a promising direction worth exploring. On the other hand, Python is often not supported for data queries in numerous real-world database applications. In such instances, SQL code is essential, rendering the PiP approach inapplicable.
3) By also considering the complementary results from Table~\ref{tab:app-code-performance}, we note that on simple and moderate tasks, PIS outperforms PiP. This indicates that SQL may have certain advantages over Python in handling straightforward data query tasks. We believe that combining the respective strengths of SQL and Python is a promising future research direction. 

To analyze the contribution of different Python generation strategies to the final SQL program, we present the distribution of Python strategy selection rates in Table~\ref{tab:method-distribution}. A strategy is considered selected when the final SQL program is guided by the Python program generated in that specific strategy. We find that all three strategies contribute to the final SQL program, with the merge strategy accounting for the largest share. Additionally, higher-performing backbones tend to exhibit a more balanced contribution across various strategies, possibly because they can generate effectively across various strategies, whereas weaker LLMs rely more on appropriate prompt guidance.

\section{Conclusion}
In this paper, we present \mname, a high-resource programming language-guided SQL generation system with two key stages: intermediate guidance preparation and guided SQL generation. Experiments across various benchmarks and difficulty levels prove that our zero-shot method, \mname, significantly outperforms all baselines, including those with few-shot examples or requiring fine-tuning. The success of \mname underscores the potential of leveraging programming languages as an intermediate step in guiding code generation, offering new insights for future text-to-SQL research.


\section*{Limitations}
\mname depends on multiple generated Python codes to guide the LLM in generating SQL programs. 
This process introduces additional inference tokens, leading to higher computational costs during test time. One potential solution to alleviate this issue is to introduce a router that selectively schedules text queries for either direct generation or Python-guided generation. We plan to explore this approach in future work.

\bibliography{custom}

\appendix

\section{More Experiment Details} \label{sec:app_exp_details}

\subsection{Benchmark Details} \label{sec:app_benchmark}
We conduct experiments on two widely recognized open-sourced text-to-SQL datasets: BIRD \cite{bird} and Archer \citep{zheng-etal-2024-archer}.  
BIRD contains over 12,751 unique question-SQL pairs derived from 95 large-scale databases spanning over 37 professional domains. The databases are designed to mimic real-world scenarios, featuring messy data rows and complex schemas. 
Archer is a human-labeled dataset focused on text-to-SQL queries involving Arithmetic, Commonsense, and Hypothetical Reasoning. We use its English subset with 1,042 question-SQL pairs, spanning 20 English-language databases across 20 distinct domains.

\subsection{Baselines} \label{sec:app_baselines}
We compare \mname with ten baselines. 

\noindent $\bullet$ \textbf{Vanilla}: This method is the same as \mname, except without Python guidance.

\noindent $\bullet$ \textbf{C3}~\citep{c3}: C3 is a zero-shot text-to-SQL method that incorporates Clear Prompting, Calibration with Hints, and Consistent Output to optimize model input, mitigate biases, and maintain output consistency, respectively.

\noindent $\bullet$ \textbf{DIN-SQL}~\citep{dinsql}: DIN-SQL tackles the text-to-SQL task by decomposing it into smaller, manageable sub-tasks, solving them in an adaptive in-context learning framework that adjusts based on the task at hand.

\noindent $\bullet$ \textbf{DAIL-SQL}~\citep{gao2024text}: DAIL-SQL uses code prompts to represent the query and selects in-context learning examples based on the query and its pre-generated SQL.

\noindent $\bullet$ \textbf{TA-SQL}~\citep{tasql}: TA-SQL leverages schema linking and logical synthesis alignment modules, in conjunction with in-context learning, to mitigate hallucinations.

\noindent $\bullet$ $\mathbf{R}^{\mathbf{3}}$~\citep{tasql}: $R^3$ establishes a framework with direct Python guidance and consensus-based refinement for text-to-SQL tasks.

\noindent $\bullet$ \textbf{CHESS}~\citep{chess}: CHESS is a multi-agent framework for text-to-SQL using in-context learning, consisting of agents such as the Information Retriever, Schema Selector, Candidate Generator, and Unit Tester. For a fair comparison, we exclude the Unit Tester agent.

\noindent $\bullet$ \textbf{CHASE-SQL}~\citep{chasesql}: CHASE-SQL enhances text-to-SQL performance by utilizing a divide-and-conquer generation approach, chain-of-thought reasoning for refinement, and instance-aware synthetic few-shot example generation. Additionally, it trains a candidate selection model using the BIRD training set. 

\noindent $\bullet$ \textbf{E-SQL}~\citep{esql}: E-SQL leverages direct schema linking via question enrichment and incorporates candidate predicates to address key challenges in text-to-SQL, including complex schemas, query ambiguity, and intricate SQL generation.

\noindent $\bullet$ \textbf{RSL-SQL}~\citep{cao2024rsl}: RSL-SQL combines techniques including bidirectional schema linking, contextual information augmentation, binary selection strategy, and multi-turn self-correction, achieving robust schema linking and thus improving text-to-SQL performance.

\subsection{Data License and Usage}
The BIRD dataset is licensed under CC BY-SA 4.0, and the Archer dataset is licensed under CC BY 4.0. Both datasets are used exclusively for academic research purposes following their respective licenses.

\section{More Experiment Results}
\subsection{Inference Cost Comparision}
\label{app:cost_compare}

Table~\ref{tab:token_usage} and Table~\ref{tab:token_cost} quantify the average inference cost per query on the BIRD dev set across different methods. While \mname incurs higher computational costs on the BIRD dev set overall than vanilla (by 6.136\$), C3 (by 6.485\$), DAIL-SQL (by 3.300\$), TA-SQL (by 4.178\$), and RSL-SQL (by 2.025\$), it remains less than DIN-SQL (by 4.421\$), CHESS (by 10.608\$), R\textsuperscript{3} (by 2.966\$), E-SQL (by 2.854\$), and CHASE-SQL (by 1.108\$). Critically, this modest cost increase is justified by \mname's significant performance gains over all baselines (see Table~\ref{tab:main-results}), demonstrating its practical efficiency for real-world deployment.

\begin{table*}[htbp]
\centering
\resizebox{0.8\textwidth}{!}{%
\begin{tabular}{lcccccccc}
\toprule
\multirow{2}{*}{Method} & \multicolumn{2}{c}{Overall} & \multicolumn{2}{c}{Simple} & \multicolumn{2}{c}{Moderate} & \multicolumn{2}{c}{Challenging} \\
\cmidrule(lr){2-3} \cmidrule(lr){4-5} \cmidrule(lr){6-7} \cmidrule(lr){8-9}
 & Input Avg. & Output Avg. & Input Avg. & Output Avg. & Input Avg. & Output Avg. & Input Avg. & Output Avg. \\
 \midrule
Vanilla & 1005 & 272 & 950 & 244 & 1085 & 306 & 1094 & 341 \\  
C3 & 529 & 138 & 524 & 111 & 536 & 161 & 537 & 235 \\  
DIN-SQL & 28185 & 1124 & 23555 & 905 & 37073 & 1422 & 29275 & 1569 \\  
DAIL-SQL & 3153 & 1789 & 2988 & 1686 & 3364 & 1939 & 3530 & 1968 \\  
TA-SQL & 6917 & 212 & 7446 & 195 & 6192 & 221 & 5859 & 288 \\
CHESS & 42573 & 2008 & 41256 & 1683 & 44623 & 2397 & 44416 & 2837 \\
R\textsuperscript{3} & 14578 & 3472 & 12970 & 3656 & 16687 & 3109 & 18092 & 3465 \\
E-SQL & 25335 & 702 & 25996 & 816 & 24003 & 886 & 25409 & 754 \\
CHASE-SQL & 12878 & 2551 & 12080 & 1225 & 13745 & 1244 & 15202 & 1752 \\
RSL-SQL & 12207 & 450 & 12518 & 454 & 11849 & 434 & 11461 & 470 \\
\midrule
\mname & 8117 & 2938 & 7666 & 2734 & 8744 & 3191 & 8987 & 3431 \\  
\bottomrule
\end{tabular}}
\caption{Inference token usage of different methods in average on BIRD dev set.}
\label{tab:token_usage}
\end{table*}

\begin{table*}[htbp]
\centering
\resizebox{0.8\textwidth}{!}{%
\begin{tabular}{lcccccccc}
\toprule
\multirow{2}{*}{Method} & \multicolumn{2}{c}{Overall} & \multicolumn{2}{c}{Simple} & \multicolumn{2}{c}{Moderate} & \multicolumn{2}{c}{Challenging} \\
\cmidrule(lr){2-3} \cmidrule(lr){4-5} \cmidrule(lr){6-7} \cmidrule(lr){8-9}
 & Input (\$) & Output (\$) & Input (\$) & Output (\$) & Input (\$) & Output (\$) & Input (\$) & Output (\$) \\
\midrule

Vanilla & 0.347 & 0.376 & 0.198 & 0.204 & 0.113 & 0.128 & 0.036 & 0.045 \\  
C3 & 0.183 & 0.191 & 0.109 & 0.093 & 0.056 & 0.067 & 0.018 & 0.031 \\  
DIN-SQL & 9.728 & 1.552 & 4.902 & 0.754 & 3.870 & 0.594 & 0.955 & 0.205 \\  
DAIL-SQL & 1.088 & 2.471 & 0.622 & 1.404 & 0.351 & 0.810 & 0.115 & 0.257 \\  
TA-SQL & 2.388 & 0.293 & 1.550 & 0.163 & 0.647 & 0.093 & 0.191 & 0.038 \\  
CHESS & 14.694 & 2.773 & 8.586 & 1.401 & 4.659 & 1.001 & 1.449 & 0.370 \\
R\textsuperscript{3} & 5.032&4.793&2.699&3.044&1.742&1.298&0.590&0.452 \\
E-SQL & 8.744&0.969&5.410&0.679&2.506&0.370&0.829&0.098\\
CHASE-SQL & 4.445&3.522&2.514&1.020&1.435&0.519&0.496&0.229\\
RSL-SQL & 4.213&0.621&2.605&0.378&1.237&0.181&0.374&0.061\\
\midrule
\mname & 2.802 & 4.057 & 1.596 & 2.277 & 0.913 & 1.333 & 0.293 & 0.448 \\  
\bottomrule
\end{tabular}}
\caption{Inference cost (GPT-4o-mini) of different methods for solving all queries of BIRD dev set (\$).}
\label{tab:token_cost}
\end{table*}

\subsection{Case Study for Code Generation} \label{app:case_code_generation}
\begin{figure*}[htbp]
\centering
\includegraphics[width=0.9\textwidth]{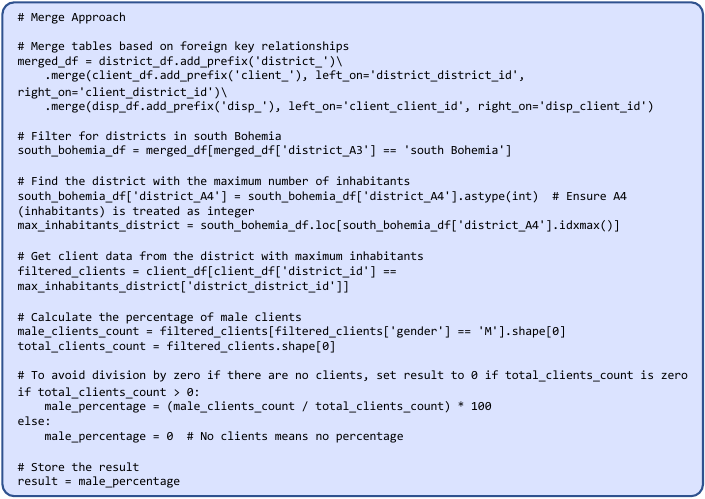}
\caption{A merge code generation case. The merge code merges all tables first, filters districts in "South Bohemia," and calculates metrics from the joined dataset.}
\label{fig:merge-approach}
\end{figure*}

\begin{figure*}[htbp]
\centering
\includegraphics[width=0.9\textwidth]{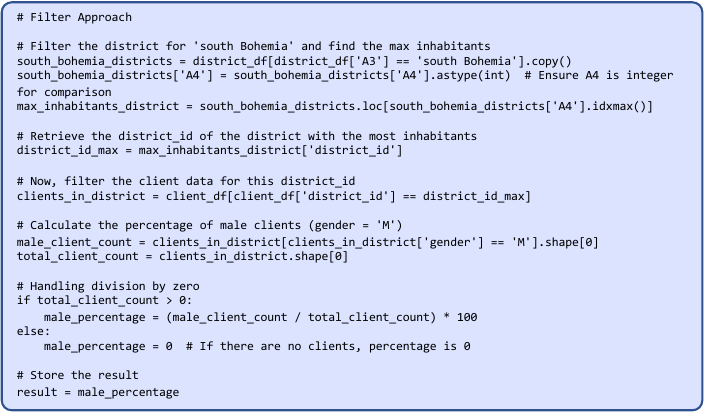}
\caption{A filter code generation case. The filter code filters districts and isolates the target district, and then filters clients, emphasizing modularity and memory efficiency.}
\label{fig:filter-approach}
\end{figure*}

\begin{figure*}[htbp]
\centering
\includegraphics[width=0.9\textwidth]{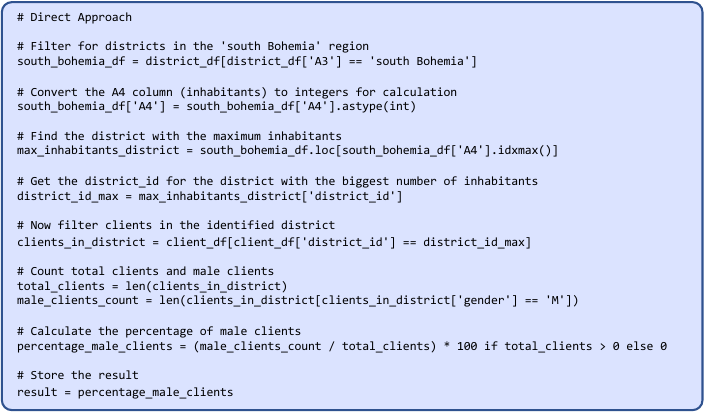}
\caption{A direct code generation case. The direct code filters districts in "south Bohemia," identifies the most populous district, and then queries client data using the retrieved district\_id.}
\label{fig:direct-approach}
\end{figure*}

We present a comparative analysis in Figure~\ref{fig:merge-approach}, Figure~\ref{fig:filter-approach}, and Figure~\ref{fig:direct-approach}  to further demonstrate the necessity of incorporating these code generation methods. These figures illustrate query execution strategies and code generation outputs across three methods: direct, merge, and filter. Direct code employs sequential task execution without table merging, instead explicitly linking data through foreign keys. This method prioritizes simplicity and transparency. Merge code first consolidates tables into a joined dataset before computing metrics, favoring holistic data integration. Filter code optimizes efficiency by eliminating irrelevant data at the early stages of processing. These approaches exhibit distinct characteristics, with each proving optimal under specific problem constraints. For instance, direct code excels in straightforward scenarios requiring traceability, while merge code suits complex multi-table analyses, and filter code benefits resource-intensive tasks. This observation aligns with the conclusions in Section~\ref{sec:diff_backbone}, where we analyze method selection patterns across varying difficulty levels and LLM backbones.

\subsection{Complementary Python Programs performance}
We provide additional performance results of the Python programs on the BIRD dev set at various difficulty levels in Table~\ref{tab:app-code-performance}.

\begin{table*}[htbp]
\centering
\resizebox{0.8\textwidth}{!}{%
\begin{tabular}{lcccccccccccc}
\toprule
\multirow{2}{*}{Model} & \multicolumn{3}{c}{Overall} & \multicolumn{3}{c}{Simple} & \multicolumn{3}{c}{Moderate} & \multicolumn{3}{c}{Challenging} \\
\cmidrule(lr){2-4} \cmidrule(lr){5-7} \cmidrule(lr){8-10} \cmidrule(lr){11-13}
 & VanS&  {\sc Pi}S & {\sc Pi}P & VanS&  {\sc Pi}S & {\sc Pi}P  & VanS&  {\sc Pi}S & {\sc Pi}P & VanS&  {\sc Pi}S & {\sc Pi}P \\
\midrule
Qwen-Coder   & 59.97 & 67.40& 66.42 & 64.75 & 72.86 & 70.70 & 55.17 & 59.91 &59.48 & 44.82 & 56.55 & 61.37 \\
QwQ & 55.08 & 64.34 & 65.58 & 62.48 & 71.13 & 69.08 & 46.33 & 56.89 &55.38 & 35.86 & 44.82 & 62.06 \\
Gemma3  & 58.86 & 66.55 & 62.12 & 64.32 & 72.75 & 67.24 & 51.72 & 57.11 & 54.31 & 46.89 & 57.24 & 54.48 \\
GPT-4o-mini & 55.35 &64.54  & 65.44 & 60.86 &70.92 & 70.27 & 48.28 & 56.47  & 57.75 & 42.76 & 49.66 & 59.31\\
\bottomrule
\end{tabular}}
\caption{Performance of Python program on the BIRD dev set. VanS and {\sc Pi}S denote SQL performance without and with Python guidance, while {\sc Pi}P denotes Python performance.}
\label{tab:app-code-performance}
\end{table*}

\subsection{Complementary Contribution of Different Python Generation Strategies}
We present the distribution of Python strategy selections on the BIRD dev set across various difficulty levels in Table~\ref{tab:app-method-distribution}.

\begin{table*}[htbp]
\centering
\resizebox{0.8\textwidth}{!}{%
\begin{tabular}{lccc|ccc|ccc|ccc}
\toprule
\multirow{2}{*}{Model} & \multicolumn{3}{c|}{Overall} & \multicolumn{3}{c|}{Simple} & \multicolumn{3}{c|}{Moderate} & \multicolumn{3}{c}{Challenging} \\
\cmidrule(lr){2-4} \cmidrule(lr){5-7} \cmidrule(lr){8-10} \cmidrule(lr){11-13}
 & Merge & Filter & Direct & Merge & Filter & Direct & Merge & Filter & Direct & Merge & Filter & Direct \\
\midrule
Qwen-Coder & 34.49&35.27&30.25&35.35&34.49&30.16&33.62&37.50&28.88&31.72&33.10&35.17 \\
QwQ & 34.49&31.42&34.09&35.03&32.32&32.65&33.84&32.54&33.62&33.10&22.07&44.83 \\
Gemma3 & 31.68&33.90&34.42&31.57&34.92&33.51&32.54&31.47&35.99&29.66&35.17&35.17 \\
GPT-4o-mini & 31.94&34.16&33.90&31.14&35.46&33.41&32.54&32.54&34.91&35.17&31.03&33.79 \\
\bottomrule
\end{tabular}}
\caption{Distribution of Python selection rates across different Python generation strategies. The percentages represent the proportion of SQL outputs generated by each strategy (merge, filter, or direct) that are selected as the final answer, summing to 100\% for each model across all strategies.}
\label{tab:app-method-distribution}
\end{table*}

\subsection{Comparison with Fine-tuned Methods} 
\label{app:compare_finetuned}
In this section, we compare \mname to methods that are explicitly fine-tuned on the text-to-SQL task. Specifically, we compare with CodeS \cite{codes} and RESD-SQL \cite{li2023resdsql}. CodeS are state-of-the-art fully open-source language models (1B–15B parameters) designed for the text-to-SQL task. It employs incremental pretraining on a curated SQL-centric corpus, enhancing SQL generation, schema linking, and domain adaptation via strategic prompt construction and bi-directional data augmentation. RESD-SQL is fine-tuned from T5 series models on the Spider training set, with explicitly designed model structure to decouple schema linking from skeleton parsing.

We find in Table \ref{app:analysCompareToFinetune} that \mname consistently outperforms the fine-tuned baselines by 6.04 to 31.44 EX and 19.59 to 47.08 VES overall, demonstrating its superiority in terms of both accuracy and efficiency. The performance gap is more pronounced in the moderate and challenging level subsets, indicating that guiding with a high-resource programming language is more effective than fine-tuning for solving difficult text-to-SQL problems. The performance of fine-tuned models heavily depends on the quality and quantity of the fine-tuning data. However, creating a large-scale dataset with challenging text-to-SQL pairs is costly and difficult. Conversely, \mname leverages Python to guide LLMs in handling challenging SQL programs without the need for a high-quality fine-tuning dataset.

\begin{table}[t]
    \centering
    \resizebox{\columnwidth}{!}{%
    \begin{tabular}{lcccccccc}
    \toprule
    \multirow{2}{*}{Method}& \multicolumn{2}{c}{Overall} & \multicolumn{2}{c}{Simple} & \multicolumn{2}{c}{Moderate} & \multicolumn{2}{c}{Challenging}  \\
    \cmidrule(lr){2-3} \cmidrule(lr){4-5} \cmidrule(lr){6-7} \cmidrule(lr){8-9}
     & EX & VES & EX & VES & EX & VES & EX & VES  \\ \midrule
    SFT CodeS-1B & 50.30 & 52.45 & 58.70 & 61.11 & 37.60 & 39.89 & 36.80 & 37.38 \\
    SFT CodeS-3B & 54.90 & 58.28 & 62.80 & 64.96 & 44.30 & 50.98 & 38.20 & 38.99 \\
    SFT CodeS-7B & 57.00 & 60.83 & 64.60 & 66.88 & 46.90 & 49.53 & 40.30 & 58.42 \\
    SFT CodeS-15B & 58.50 & 61.54 & 65.80 & 67.87 & 48.80 & 51.69 & 42.40 & 52.71 \\
    RESDSQL-Base & 33.10 & 34.05 & 42.30 & 42.75 & 20.20 & 22.16 & 16.00 & 16.54 \\
    RESDSQL-Large & 38.60 & 40.81 & 46.50 & 47.21 & 27.70 & 30.00 & 22.90 & 34.67 \\
    RESDSQL-3B & 43.90 & 45.64 & 53.50 & 53.35 & 33.30 & 35.49 & 16.70 & 28.84
    \\\midrule
    \mname &\textbf{64.54}& \textbf{81.13}& \textbf{70.92}& \textbf{88.88}& \textbf{56.47}& \textbf{71.06}& \textbf{49.66}& \textbf{63.92} \\ 
    \bottomrule
    \end{tabular}
    }
        \caption{Comparison with fine-tuned models on the BIRD dev set. The baseline results are cited from~\citet{li2024dawn}. Since the baselines exclusively report VES, this table presents VES scores rather than R-VES. The best result for each case is highlighted in \textbf{bold}.}
        \label{app:analysCompareToFinetune}
\end{table}

\subsection{Case Study} \label{app:casestudy}
\begin{figure}[t]
    \centering
    \includegraphics[width=1.0\columnwidth]{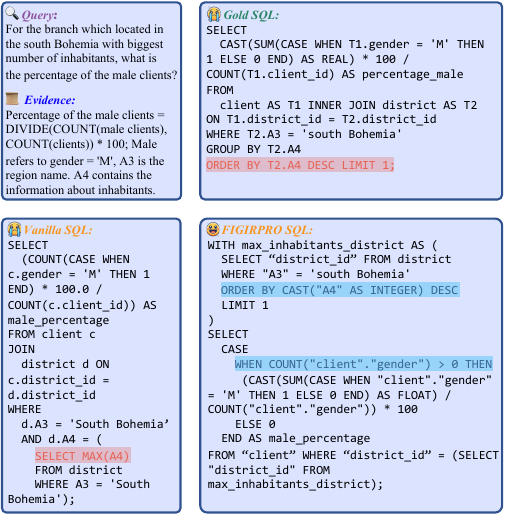}
    \caption{Case study of a specific query and its corresponding evidence, showcasing the gold SQL and the SQL generated by the vanilla method and \mname.}
    \label{fig:app_case_sql}
    \vspace{-10pt}
\end{figure}

We conduct a case study in Figure~\ref{fig:app_case_sql}, comparing the SQL program generated by \mname with both the gold SQL program and the one generated by the vanilla method. Both the gold SQL and vanilla return incorrect SQL programs as they directly sort by \texttt{A4}, which is of text type. In contrast, \mname generates the correct SQL program by casting \texttt{A4} to an integer type before sorting. The SQL program generated by \mname is also superior to the other two in terms of robustness, execution efficiency, and readability. \mname is more robust as it incorporates comprehensive considerations in case statements, such as avoiding division-by-zero errors by checking \texttt{COUNT("client"."gender") > 0}. It is more efficient and readable as it structurally decouples the identification of the target district (the most populous district in South Bohemia) from the subsequent calculation of male client percentages. In contrast, the gold SQL and the vanilla SQL combine filtering, table joins, grouping, and calculations into a single monolithic block, resulting in low execution efficiency and poor readability.

\subsection{Execution time of generated code}
\label{app:execution_time}
This section details the average execution time for the Python and SQL code generated by \mname (as shown in Table~\ref{app:analysTimeCost}). At approximately 1 second per query, this execution time is negligible compared to the inference cost.

\begin{table}[t]
    \centering
    \resizebox{1.0\columnwidth}{!}{%
    \begin{tabular}{lcccccccc}
    \toprule
    \multirow{2}{*}{Method}& \multicolumn{2}{c}{Overall} & \multicolumn{2}{c}{Simple} & \multicolumn{2}{c}{Moderate} & \multicolumn{2}{c}{Challenging}  \\
    \cmidrule(lr){2-3} \cmidrule(lr){4-5} \cmidrule(lr){6-7} \cmidrule(lr){8-9}
     & Python & SQL & Python  & SQL & Python  & SQL & Python  & SQL  \\ \midrule
    \mname &0.961& 0.166& 0.905& 0.170& 1.036& 0.170& 1.082& 0.124 \\ 
    \bottomrule
    \end{tabular}
    }
        \caption{Execution time (s) of the generated code (average per query).}
        \label{app:analysTimeCost}
        \vspace{-10pt}
\end{table}

\subsection{Combination with refinement} \label{sec:app_simple_refinement}
To demonstrate that \mname can be readily integrated with refinement techniques, we present its performance in Table~\ref{tab:refinement} using a simple refinement strategy. Specifically, if no generated SQL query produces results that align with the Python-voted outcome, the SQL query is regenerated.

\begin{table}[t]
    \centering
    \resizebox{\columnwidth}{!}{%
    \begin{tabular}{lcccccccc}
    \toprule
    \multirow{2}{*}{Method}& \multicolumn{2}{c}{Overall} & \multicolumn{2}{c}{Simple} & \multicolumn{2}{c}{Moderate} & \multicolumn{2}{c}{Challenging}  \\
    \cmidrule(lr){2-3} \cmidrule(lr){4-5} \cmidrule(lr){6-7} \cmidrule(lr){8-9}
     & EX & R-VES & EX & R-VES & EX & R-VES & EX & R-VES  \\ \midrule
    \mname &64.54& 63.71& 70.92& 70.06& 56.47& 55.63& 49.66& 49.06 \\ 
    +refinement &\textbf{65.12}& \textbf{66.85}& \textbf{71.03}& \textbf{73.01}& \textbf{56.90}& \textbf{58.17}& \textbf{53.79}& \textbf{55.40} \\ 
    \bottomrule
    \end{tabular}
    }
        \caption{The results of \mname with a simple refinement strategy.}
        \label{tab:refinement}
\end{table}

\subsection{Scaling experiments} \label{app:scale}
To evaluate the generality and robustness of \mname across models of different scales, we also conducted experiments using the Qwen2.5-Coder-Instruct series across different model sizes in Table~\ref{tab:model-size}. We observe that \mname consistently improves both EX and R-VES scores across models. Overall, the performance gains become more pronounced with larger models, likely due to their stronger Python reasoning and generalization capabilities. This reveals the potential of our method when applied to more powerful language models. These results further demonstrate the effectiveness of our approach.
\begin{table}[t]
\centering
\resizebox{1.0\columnwidth}{!}{%
\begin{tabular}{lcccccccc}
\toprule
\multirow{2}{*}{\textbf{Model}} & \multicolumn{2}{c}{Overall} & \multicolumn{2}{c}{Simple} & \multicolumn{2}{c}{Moderate} & \multicolumn{2}{c}{Challenging} \\
\cmidrule(lr){2-3} \cmidrule(lr){4-5} \cmidrule(lr){6-7} \cmidrule(lr){8-9}
 & EX & R-VES & EX & R-VES & EX & R-VES & EX & R-VES \\
 \midrule
\multicolumn{9}{l}{\textit{Vanilla method}} \\
\midrule
Qwen2.5-Coder-7B &51.63&49.76&60.97&59.10&39.66&37.55&30.34&29.27 \\
Qwen2.5-Coder-14B &61.21&58.20&68.43&64.95&51.94&49.72&44.83&42.31 \\
Qwen2.5-Coder-32B &59.97&58.61&64.76&63.64&55.17&53.40&44.83&43.12 \\
\midrule
\multicolumn{9}{l}{\textit{With \mname}} \\
\midrule
Qwen2.5-Coder-7B &54.24&56.02&62.81&64.93&42.24&43.46&37.93&39.33 \\
Qwen2.5-Coder-14B &65.45&67.77&71.78&74.41&56.47&58.19&53.79&56.07 \\
Qwen2.5-Coder-32B  & 67.40 & 65.54 & 72.86 & 71.14 & 59.91 & 57.89 & 56.55 & 54.29 \\
\bottomrule
\end{tabular}}
\caption{Performance of \mname with LLMs of different scales on the BIRD dev set. The best result for each case is highlighted in \textbf{bold}.}
\label{tab:model-size}
\vspace{-10pt}
\end{table}

\section{Broader Impacts}
Our \mname method allows non-technical users to generate SQL queries using natural language, improving productivity and making data more accessible. It can benefit fields like healthcare, finance, and education by enabling faster, data-driven decisions without requiring SQL expertise. However, the method could also be misused to query leaked or unauthorized databases, risking privacy breaches. To address this, robust access controls and privacy safeguards must be implemented to ensure responsible use.
\end{document}